\documentclass[12pt, final]{l4dc2022}

\title[Distributed Control using RL with Temporal-Logic-Based Reward Shaping]{Distributed Control using Reinforcement Learning with Temporal-Logic-Based Reward Shaping}
\usepackage{times}
\usepackage{comment}
\usepackage{graphics}

 \author{\Name{Ningyuan Zhang} \Email{ningyuan@bu.edu}\\
  \Name{Wenliang Liu} \Email{wliu97@bu.edu}\\
  \Name{Calin Belta} \Email{cbelta@bu.edu}\\
  \addr Boston University, Massachusetts, USA}

\newtheorem{problem}{Problem}[section]

\begin{document}

\maketitle

\begin{abstract}%
We present a computational framework for synthesis of distributed control strategies for a heterogeneous team of robots in a partially observable environment. The goal is to cooperatively satisfy specifications given as Truncated Linear Temporal Logic (TLTL) formulas. Our approach formulates the synthesis problem as a stochastic game and employs a policy graph method to find a control strategy with memory for each agent. We construct the stochastic game on the product between the team transition system and a finite state automaton (FSA) that tracks the satisfaction of the TLTL formula. We use the quantitative semantics of TLTL as the reward of the game, and further reshape it using the FSA to guide and accelerate the learning process. Simulation results demonstrate the efficacy of the proposed solution under demanding task specifications and the effectiveness of reward shaping in significantly accelerating the speed of learning. 
\end{abstract}


\section{Introduction}
\label{sec:intro}
Many recent works treat the sequential decision-making problem of multiple autonomous, interacting agents as a Multi-Agent Reinforcement Learning (MARL) problem. In this framework, each agent learns a policy to optimize a long-term discounted reward through interacting with the environment and the other agents \cite{zhang2021multi, qie2019joint, cui2019multi}. In many realistic applications, the environment states are only partially observable to the agents \cite{hausknecht2015deep}. Learning in such environments has become a recurrent problem in MARL \cite{zhang2021multi}, the difficulty of which lies in the fact that the optimal policy might require a complete history of the whole system to determine the next action to perform \cite{meuleau1999learning}. The authors in \cite{peshkin2001learning} used finite policy graphs to represent policies with memory and successfully applied a policy gradient MARL algorithm in a fully cooperative setting. 

In this paper, we consider the problem of controlling a heterogeneous team of agents to satisfy a specification given as a Truncated Linear Temporal Logic (TLTL) formula \cite{li2017reinforcement} using MARL. TLTL is a version of Linear Temporal Logic (LTL) \cite{pnueli1977temporal}. The semantics of TLTL is given over a finite trajectory in a set such as the state space of the system. TLTL has dual semantics. In its Boolean (qualitative) semantics (whether the formula is satisfied), a trajectory satisfying a TLTL formula is accepted by a Finite State Automaton (FSA). The quantitative semantics assigns a degree of satisfaction (robustness) of a formula. Designing a reward function that accurately represents the specification is an important issue in reinforcement learning. We use the robustness of the TLTL specification as the reward of the MARL problem. However, the TLTL robustness is evaluated over the entire trajectory, so we can only get the reward at the end of each episode. To address this, we propose a novel reward shaping technique that introduces two additional reward terms based on the quantitative semantics of TLTL and the corresponding FSA. Such rewards, which are obtained after each step, guide and accelerate the learning. 

As in \cite{peshkin2001learning}, we consider a partially observable environment and train a finite policy graph for each agent. When deploying the policy, each agent only knows its own state. However, during training we assume all agents know the states of each other. Then we use this information to guide and accelerate learning. This assumption is reasonable in practice because the training environment is highly configurable and a communication system can be easily constructed, while during deployment there is no guarantee that such a communication system is available. 

The idea of modifying rewards using the semantics of temporal logics was introduced in \cite{li2017reinforcement} to address the problem of single agent learning in a fully observable environment. The method was then extended to deep RL to successfully control two manipulator arms working together to make and serve hotdogs \cite{li2019formal}. Although the problem in \cite{li2019formal} involved two agents, single agent RL algorithms were used with two different specially designed task specifications. 
In \cite{sun2020automata}, temporal logic rewards were used for multi-agent systems. As opposed to our work, the environment in \cite{sun2020automata} is fully observable. Moreover, each sub-task in \cite{sun2020automata} is restricted to be executed by one agent. In this paper, we allow both independent tasks, which can be accomplished by one agent, and shared tasks, which must be achieved by cooperation of several agents. The rewards in \cite{li2019formal} and \cite{sun2020automata} encourage the FSA to leave its current state, without distinguishing whether it is a transition towards the satisfaction of the TLTL specification. In this paper, the combination of the two reshaped rewards encourage transitions towards satisfaction. Another related work is \cite{hammond2021multi}, in which the authors apply an actor-critic algorithm to find a control policy for a multi-agent system that maximizes the probability of satisfying an LTL specification. However, the reward does not capture a quantitative semantics, and a fully observable environment is assumed. 

The contributions of this paper can be summarized as follows. First, we propose a general procedure that uses MARL algorithms to synthesize distributed control from arbitrary TLTL specifications for a heterogeneous team of agents. Each agent can have different capabilities and both independent and shared sub-tasks can be defined. The agents work in a partially observable environment, and the policy for each agent only requires its own state during deploying. During training, we assume a fully observable environment, which enables the use of TLTL robustness as a reward. Second, we create a novel temporal logic reward shaping technique, where two additional rewards based on TLTL robustness and FSA states are added. Both rewards are obtained immediately after each step, which guides and accelerates the learning process.

\section{Preliminaries and Notation}
\label{sec:Preliminaries}
Given a set $S$, we use $|S|$, $2^S$, and $S^*$ to denote its cardinality, the set of all subsets of $S$, and a finite sequence over $S$. Given a set $\Sigma$, a collection of sets $\Delta = \{\Sigma_i \subset \Sigma, i\in I\}$, in which $I$ is a set of labels, is called a distribution of $\Sigma$ if $\cup_{i\in I} \Sigma_i = \Sigma$. For $\sigma \in \Sigma$, we use $I_{\sigma} = \{i \in I | \sigma \in \Sigma_i\}$ to denote the set of labels of elements in $\Delta$ that contains $\sigma$.

\subsection{Partially Observable Stochastic Games}\label{sec:POSG}

Single agent reinforcement learning uses Markov Decision Processes (MDP) as mathematical models \cite{puterman2014markov}. In MARL, more complicated models are needed to describe the interactions between the agents and the environment. A popular model is a Stochastic Game (SG) \cite{bucsoniu2010multi}. A Partially Observable Stochastic Game (POSG), which is a generalization of an SG, is defined as a tuple $\langle I, S, d_0, \{A_i,O_i,B_i\}_{i\in I},  p, \{r_i\}_{i\in I}\rangle$, in which $I=\{1,\ldots,n\}$ is an index set for the agents; $S$ is the discrete joint state space; $d_0$ is the probability distribution over initial states; $A_i$ is the set of actions for agent $i$, $O_i$ is the discrete observation space and $B_i: S \to O_i$ is the observation function; $p: S \times A \to Pr(S)$ is the transition function that maps the states of the game and the joint action of the agents, defined as $A = \times_{i=1}^n A_i$, to probability distributions over states of the game at next time step; and $r_i: S \times A \times S \to \mathbb{R}$ is the reward function for agent $i$. If every agent is able to obtain the complete information of the environment state when making decisions, i.e., $O_i = S$ and $B_i(s) = s$ for all $s \in S$, then the POSG becomes a fully observable SG. 

The goal for each agent $i$ is to find a policy $\pi_i$ that maximizes its value $V_i$:
\begin{equation*}
    V_i(\vec{\pi}, d_0) = \sum_{t=0}^{T-1}\gamma^t \mathbb{E}_{a_t\sim\vec{\pi}}(r_i(s_t,a_t,s_{t+1})|\vec{\pi}, d_0),
\end{equation*}
where $\gamma \in (0,1]$ is the discount factor and $\vec{\pi} = (\pi_1, ..., \pi_n)$ is a set of policies. Note that for each agent the environment is non-stationary since its reward $r_i$ may depend on other agents' policies. A \emph{memory-less policy} $\pi_i:O_i\rightarrow Pr(A_i)$ is a mapping from the observations of agent $i$ to probability distributions over the action space of agent $i$. For a fully observable SG,  memory-less policies are sufficient to achieve optimal performance. However, for POSGs, the best memoryless policy can still be arbitrarily worse than the best policy using memory \cite{singh1994learning}. A policy graph \cite{meuleau1999learning} is a common way to represent a policy with memory. 

\subsection{Truncated Linear Temporal Logic}\label{sec:TLTL}
Truncated Linear Temporal Logic (TLTL) \cite{li2017reinforcement} is a predicate temporal logic inspired from the traditional Linear Temporal Logic (LTL) \cite{pnueli1977temporal}. TLTL 
can express rich task specifications that are satisfied in finite time. Its formulas 
are defined over predicates $f(x) \geq 0$, where $f: X \to \mathbb{R}$ is a function over $X$ (such as the state space of a system). Then TLTL formulas are evaluated against finite sequences over $X$.  TLTL formulas have the following syntax:
\begin{multline*}
    \phi := \top \; | \; f(x) \geq 0 \; | \; \neg \phi \; | \; \phi \wedge 
\psi \; | \; \phi \vee \psi \;|\; \phi \Rightarrow \psi \;|\; \lozenge\phi \; | \;\bigcirc \phi \; | \; \phi \mathcal{U}\psi \; |\; \phi \mathcal{T} \psi,
\end{multline*}
where $\top$ is the Boolean True. $\neg$ (negation), $\land$ (conjunction) and $\lor$ (disjunction) are Boolean connectives. $\lozenge$, $\bigcirc$, $\mathcal{U}$, and $\mathcal{T}$ are temporal operators that stand for ``eventually", ``next", ``until"  and ``then" respectively. Given a sequence over $X$, $\lozenge\phi$ (eventually) requires $\phi$ to be satisfied at some time step, $\bigcirc\phi$ (next) requires $\phi$ to be satisfied at the second time step, $\phi\mathcal U\psi$ (until) requires $\phi$ to be satisfied at each time step before $\psi$ is satisfied, $\phi\mathcal T\psi$ (then) requires $\phi$ to be satisfied at least once before $\psi$ is satisfied.  TLTL also has derived operators, e.g.,  $\square$ (finite time always) and $\Rightarrow$ (imply). 

TLTL formulas can be interpreted in a qualitative semantics or a quantitative semantics. The qualitative semantics provides a yes/no answer to the corresponding property, while the quantitative semantics generates a measure of the degree of satisfaction. Given a finite trajectory $x_{t:t+k}:= x_t x_{t+1}\ldots x_{t+k}$, the quantitative semantics of a formula $\phi$, also called \textit{robustness}, is denoted by $\rho(x_{t:t+k}, \phi)$. We refer the readers to \cite{li2017reinforcement} for detailed definitions of the TLTL qualitative and quantitative semantics.

\subsection{Finite State Automata}\label{sec:FSPA}
\begin{definition}[Finite State Predicate Automaton \cite{li2019formal}]
    A finite state predicate automaton (FSPA) corresponding to a TLTL formula $\phi$ is a tuple $\mathcal{A} = \langle Q, q_0, F, T_{r}, \Psi, \mathcal{E}, b \rangle$, in which $Q$ is a set of automaton states, $q_0 \in Q$ is the initial state, $F \subset Q$ is the set of final states, $T_r$ is the set of trap states, $\Psi$ is a set of predicate Boolean formulas, where the predicates are evaluated over $X$, $\mathcal{E} \subseteq Q \times Q$ is the set of transitions, and $b:\mathcal{E} \to \Psi$ maps transitions to formulas in $\Psi$. 
\end{definition}

The semantics of FSPA is defined over finite sequences over $X$. We refer to the elements in $\Psi$ as {\it edge guards}. A transition is enabled if the corresponding Boolean formula is satisfied. Formally, by noting that a Boolean formula is a particular case of a TLTL formula, the FSPA transitions from $q_t$ to $q_{t+1}$ at time $t$ if and only if $\rho(x_{t:t+k}, b(q_t, q_{t+1})) > 0$, where $x_{t:t+k}$ is a sequence of $X$ from $t$ to $t+k$. Note that the robustness is only evaluated at $s_t$ instead of the whole sequence. Thus we abbreviate $\rho(x_{t:t+k}, b(q_t, q_{t+1}))$ to $\rho(x_t, b(q_t, q_{t+1}))$.


A trajectory over $Q$ is accepting if it ends in the final states $F$. A trajectory over $X$ is accepted by $\mathcal A$ if it leads to an accepting trajectory over $Q$.
Each TLTL can be translated into an equivalent FSPA \cite{li2019formal}, in the sense that a trajectory over $X$ satisfies the TLTL specification if and only if the same trajectory is accepted by the FSPA. All trajectories over $X$ that violate the TLTL formula drive the FSPA to the trap states in $T_r$. 


\section{Problem Formulation} \label{sec:PF}
The geometry of the world is modeled by an environment graph $G = (V, E)$, where $V$ is a set of vertices and $E \subset V\times V$ is a set of edges. The motions of the agents are restricted by this graph. Assume we have a set of agents $\{i | i \in I\}$ where $I$ is a label set, and a set of service requests $\Sigma$. Let $ l: \Sigma \to 2^V$ be a function indicating the locations at which a request occurs. For a given request $\sigma \in \Sigma$, let $V_\sigma=l(\sigma)\subseteq V$ be the set of vertices where $\sigma$ occurs. The definition of $V_\sigma$ implicitly assumes that one request can occur at multiple vertices, since $V_\sigma\in2^V$. Also, multiple requests can occur at the same vertex, since $V_{\sigma_1}$ and $V_{\sigma_2}$ may have non-empty intersection for $\sigma_1\neq\sigma_2$. We use a distribution $\Delta = \{\Sigma_i \subset \Sigma, i\in I\}$ to model the agents' capabilities for completing different service requests: $\sigma \in \Sigma_i$ means that request $\sigma$ can be serviced by agent $i$. We further define $I_\sigma=\{i\in I|\sigma\in\Sigma_i\}$ as the set of agents that can service request $\sigma$. We consider two types of services: \textit{independent} and \textit{shared}. An independent request is any $\sigma \in \Sigma$ such that $|I_\sigma| = 1$. This type of request can only be serviced by agent $i$ such that $\sigma \in \Sigma_i$. A shared request $\sigma$ is defined by $|I_\sigma| > 1$, which means that servicing it requires the cooperation of all the agents that own $\sigma$.

We model the motion and actions of agent $i$ by using a deterministic finite transition system $\mathcal{T}_i = \langle V_i, {v_0}_i, A_i, \delta_i, \Sigma_i, \models_i\rangle$ where $V_i \subseteq V$ is the set of vertices that can be reached by agent $i$; ${v_0}_i \in V_i$ is the initial location; $A_i = V_i\cup\Sigma_i\cup \{\epsilon\}$ is the discrete action space; $\delta_i:V_i \times A_i \to V_i$ is a deterministic transition function; $\Pi_i = \Sigma_i \cup \{\epsilon\}$ is a proposition set; $\models_i \subseteq V \times \Pi_i$ is the satisfaction relation such that 1) $(v,\epsilon) \in \models_i$ for all $i\in I$, $v\in V_i$,  and 2) $(v,\sigma) \in \models_i$, $\sigma \in \Sigma_i$, if and only if  $v \in l(\sigma)$. The meaning of taking action $a\in A_i$ at state $v$ is as follows: if $a \in V_i$, then the agent tries to move to vertex $a$ from $v$; if $a \in \Sigma_i$, then agent $i$ tries to conduct service $a$ at $v$; $a = \epsilon$ indicates that the agent stays at the current vertex without conducting any requests. Formally,
\begin{equation}
    \delta_i(v,a) =  \begin{cases}
        v' & \text{if} \; a = v' \in V_i \;\text{and}\; (v,v') \in E \\
        v & \text{otherwise} \\
    \end{cases}
\end{equation}
For $v, v'\in V_i$ and $a \in A_i$, a transition $v' = \delta_i(v,a)$ is also denoted by $v\xrightarrow{a}_{\mathcal{T}_i} v'$. Now we define the Motion and Service (MS) plan for an agent, which is inspired by \cite{chen2011formal}. 

\begin{definition}[Motion and Service Plan] A Motion and Service (MS) plan for an agent $i \in I$ is a sequence of states represented by ordered pairs $z^i_0z^i_1 \ldots z^i_T \in (V \times \Pi_i)^*$ that satisfies the following properties:
    \begin{enumerate}
        \item $z^i_0 = ({v_0}_i,\epsilon)$.
        \item For all $t\geq0$, $z^i_t=(v,\sigma)\in\models_i$.
        \item For all $t\geq1$, given $z^i_{t-1} = (v, \sigma)$ and $z^i_t = (v', \sigma')$, if $v \neq v'$ then $\sigma' = \epsilon$.
        \item For all $t\geq1$, given $z^i_{t-1} = (v, \sigma)$ and $z^i_t = (v', \sigma')$, then $\exists a\in A_i$ such that $\delta_i(v,a) = v'$.
    \end{enumerate}
\end{definition}
We use $z^i_{0:T}$ to denote the MS plan for an agent $i$ from time $0$ to $T$, i.e., $z^i_{0:T} = z^i_0z^i_1\cdots z^i_T$. A Team Motion and Service (TMS) plan is then defined as $z_{0:T} = \times^n_{i=1}z^i_{0:T}$.

We assume that there exists a global discrete clock that synchronizes the motions and the services of requests of all agents. We also assume that the times needed to service requests are all equal to 1. In other words, similar to POSGs, at each time step, an agent either chooses to move to a vertex according to $\delta_i$ or to stay where it is to conduct a particular request in $\Sigma_i$. Before the beginning of the next time step, all motions and requests are completed and the agents are ready to execute the next state in their MS plans. A MS plan thus uniquely defines a sequence of actions for an agent. More specifically, the action derived from a MS plan $z^i$ at $t \geq 1$ is determined by $z^i_{t-1} = (v,\sigma)$ and $z^i_t = (v',\sigma')$. If $\sigma' = \epsilon$ and $v \neq v'$, then the agent moves to vertex $v$ from $v'$ (for the case where $v = v'$, the agent does not conduct any request and waits at $v$). On the other hand, $\sigma \neq \epsilon$ means that the agent should conduct request $\sigma$ and it must have reached vertex $v$ at the previous time step according to property $3$ from the above definition.
 
In this paper, we use the term \textit{global behavior} to refer to the sequence of requests serviced by the whole team. An independent request $\sigma$ is considered to be finished at time $t$ if and only if $z^i_t = (v,\sigma)$, $i \in I_\sigma$. For a shared request $\sigma$, we assume that all the agents that own this request are capable of communicating with each other at the same vertex where $\sigma$ occurs. A shared service request $\sigma$ is completed at time $t$ if and only if $z^i_t = (v_i,\sigma)$ for all $i \in I_\sigma$ and $v_j = v_i$ for all $i,j \in I_\sigma, i \neq j$. Thus, given individual MS plans of all the agents, one single global behavior is then uniquely determined. This is directly deduced from the assumption of a constant finishing time for all requests. Next we define a term called \textit{Team Trajectory} to describe the results of executing a global TMS plan.

\begin{definition}[Team Trajectory]
    Given a team of $n$ agents $\mathcal{T}_i$, a set of service requests $\Sigma$, a distribution $\Delta = \{\Sigma_i\subseteq \Sigma, i\in I\}$, a function $l: \Sigma \to V$ that shows the locations of requests and a TMS plan $z_{0:T}$ for the whole team of agents, the Team Trajectory from executing the TMS plan is a sequence of $T+1$ states $x_{0:T}$, each of which is a $(n+1)$-tuple, with the first $n$ elements equal to the vertices in the corresponding individual MS plan. The last element of the team trajectory at each time $t$ is a set $\Bar{\Sigma}_t \subseteq \Sigma$, such that $\Bar{\Sigma}_t:= \{\sigma\in\Sigma | \sigma \;\text{is finished at time t}\}$.
\end{definition}

Now the problem considered in this paper can be formulated as follows.
\begin{problem}\label{problem formulation}
    Given an environment graph $G$, a team of agents $\mathcal{T}_i$ as defined above, $i\in I$, a set of service requests $\Sigma$, a distribution $\Delta = \{\Sigma_i\subseteq \Sigma, i\in I\}$, a function $l: \Sigma \to V$ that shows locations of requests, find a set of MS plans for each agent such that the team trajectory obtained by executing individual MS plans satisfying a given global task specification encoded by a TLTL formula over predicate functions of states of the team trajectory $x_{0:T}$.
\end{problem}

A natural, classical motion planning approach to Problem \ref{problem formulation} would be to design in advance MS plans for each agent. However, this requires the transition function $\delta_i$ for each agent $i$ to be known, which is not always true in realistic applications. Moreover, a priori top-down design that guarantees satisfactory performance can become extremely difficult in complex and time-varying environments \cite{bucsoniu2010multi}. Therefore, we tackle this problem using MARL, in which agents learn how to act by constantly interacting with the environment and other agents, and adjusting their behaviors according to feedback received from the environment. The transition functions $\delta_i$ are assumed to be unknown.

\begin{figure}
\centering
\begin{minipage}[t]{0.56\textwidth}
\label{fig:setup}
\centering
\includegraphics[height=4cm]{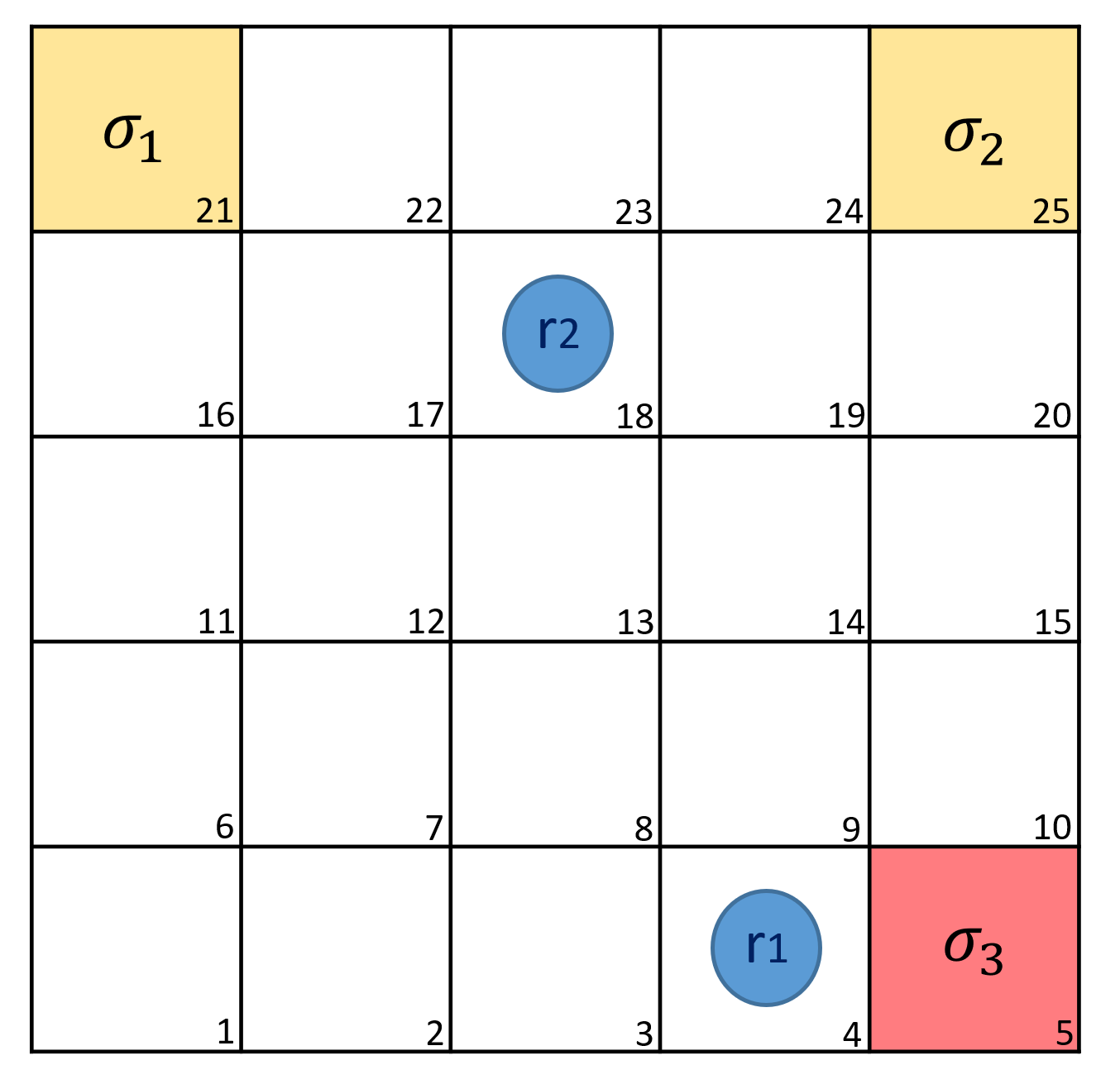}
\caption{\small The grid world of Example \ref{example}. Independent and shared service requests are depicted by yellow and red cells respectively. Cells (vertices) are labeled by the value at the bottom right corner.   }
\end{minipage}
\quad
\begin{minipage}[t]{0.40\textwidth}
\label{fig:FSPA}
\centering
\includegraphics[height=4cm]{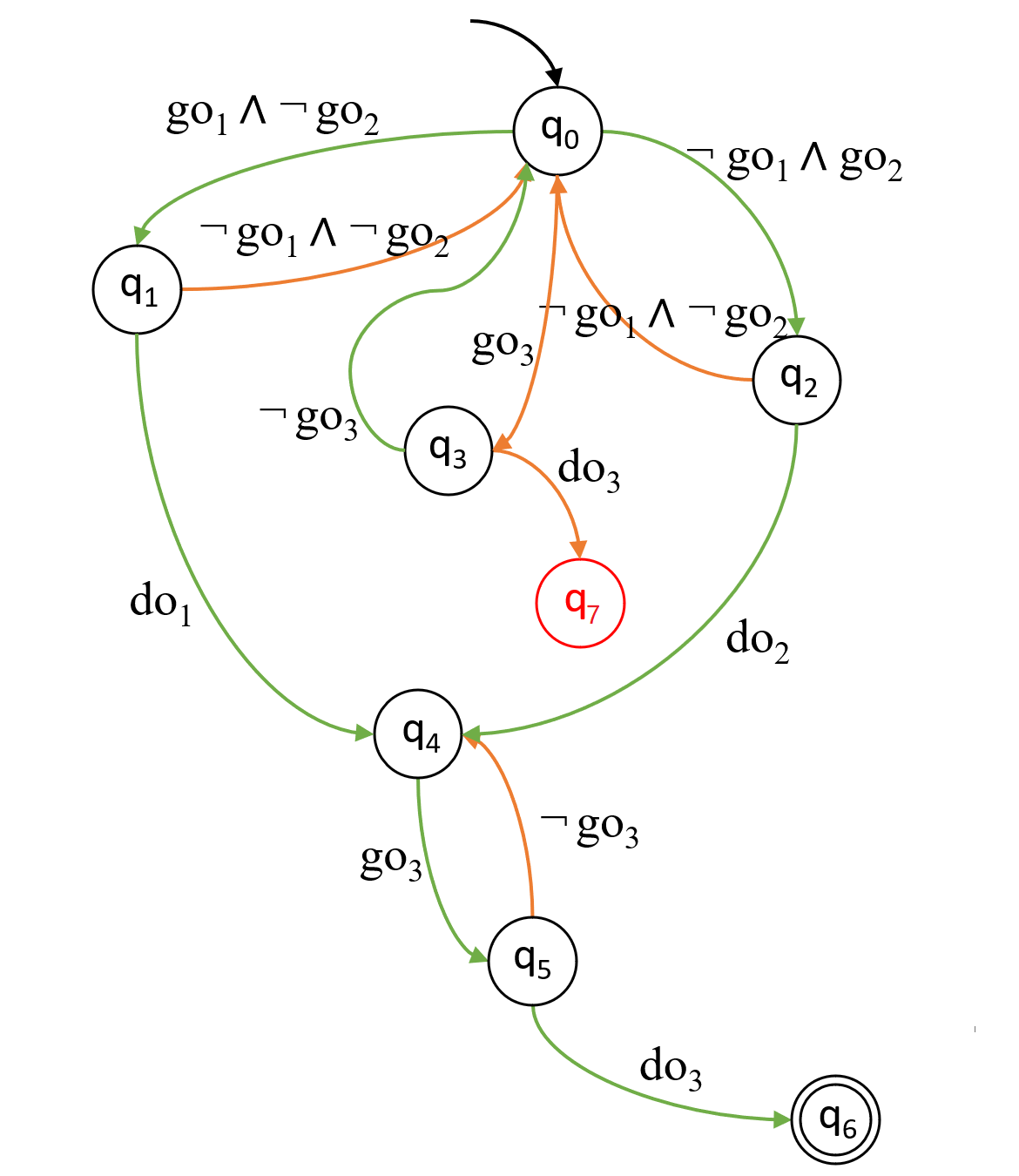}
\caption{\small The FSPA corresponding to $\phi$. State $7$ is the trap state and state $6$ is the final state.}
\end{minipage}
\vspace{-10pt}
\end{figure}

\begin{example} \label{example}
Consider a team of two heterogeneous agents that have to service three requests: $\sigma_1, \sigma_2$ and $\sigma_3$, with an additional requirement that $\sigma_3$ must not be finished until $\sigma_1$ or $\sigma_2$ has been finished. The environment is shown in Fig. \ref{fig:setup} a $5 \times 5$ as a grid world, which can be abstracted by a graph $G = (V,E)$. Each cell represents a vertex $v \in V$ (labeled with values at the bottom right) and each facet forms a reflective, two-way edge in $E$. There are three service requests: $\Sigma = \{\sigma_1, \sigma_2, \sigma_3\}$. There is a team of two robots modeled by transition systems $\mathcal{T}_1 = \langle V, v_4, A_1, \delta_1, \Sigma_1, \models_1\rangle$ and $\mathcal{T}_2 = \langle V, v_{18}, A_2, \delta_2, \Sigma_2, \models_2\rangle$. Assume both agents are able to move through any edges in $E$. Thus the action spaces are $A_1 = V\cup\Sigma_1\cup \{\epsilon\}$ and $A_2 = V\cup\Sigma_2\cup \{\epsilon\}$. The capabilities of conducting services are captured by the distribution $\Delta = \{\Sigma_1, \Sigma_2\}$, $\Sigma_1 = \{\sigma_1, \sigma_3\}$ and $\Sigma_2 = \{\sigma_2, \sigma_3\}$, which means $\sigma_1$ and $\sigma_2$ are independent requests and $\sigma_3$ is a shared request. For instance, to finish request $\sigma_1$, robot $\mathcal{T}_1$ must move to vertex $v_{21} \in l(\sigma_1)$ and then choose to take action $\sigma_1 \in A_1$. This behavior, which can be interpreted as a sub-task towards the success of the global mission, is formulated by a TLTL formula $\phi_1 = go_1(\cdot) \mathcal{T} do_1(\cdot) \land \square \neg (\neg go_1(\cdot)\land do_1(\cdot))$ where $go_1(\cdot)$ and $do_1(\cdot)$ are predicate functions defined over states of the team trajectory as
    \begin{equation}
    \begin{aligned}\label{eqn:predicates_independSigma}
        go_1(x_{t:T}) &= C_1 - \min_{v \in l(\sigma_1)} dist(v^1_t, v) \\
        do_1(x_{t:T}) &= \begin{cases}
            C_2 \quad \sigma_1 \in \bar{\Sigma}_t \\
            -C_2 \quad \text{otherwise}
        \end{cases}
    \end{aligned}
    \end{equation}
    where $v^1_t$ is the location of robot $1$ at time $t$ (included in team trajectory state $x_t$), $dist(v^1_t, l(\sigma_1))$ is the distance from $v^1_t$ to the location at which $\sigma_1$ occurs, and $C_1$ and $C_2$ are positive constants. Then the global specification is given by the TLTL formula $\phi = (\lozenge \phi_1 \lor \lozenge \phi_2) \land \lozenge \phi_3 \land \neg \phi_3 \mathcal{U} (\phi_1 \lor \phi_2)$. The FSPA corresponding to $\phi$ can be found in Fig. \ref{fig:FSPA}. The predicate functions $go_2(\cdot)$, $do_2(\cdot)$ and $do_3(\cdot)$ are defined similarly as \eqref{eqn:predicates_independSigma}. Predicates $go_3(\cdot)$ are designed in a slightly different form to encode the cooperative behaviors:
    \begin{equation}
    \label{eqn:predicates_dependSignma}
        go_3(x_{t:T}) = C_3 - \min_{v \in l(\sigma_3)}\left(\max_{i \in I_{\sigma_3}}[dist(v^i_t, v)]\right),
    \end{equation}
    where $C_3$ is some constant. Note that although formula $\phi$ is constructed hierarchically, the resultant FSPA is actually nonhierarchical.
\end{example}

\section{Problem Solution Using MARL}\label{sec:solution}
In summary, our method can be divided into three steps. First, in Sec. \ref{sec:POSG formulation}, we formulate an POSG whose solution can be easily transferred to a set of MS plans that solves Problem \ref{problem formulation}. Second, in Sec. \ref{sec:TLRS}, we introduce a reward shaping technique that employs robustness and a heuristic energy-like function to produce an additional reward signal. Finally in Sec. \ref{sec:Solving POSG}, we demonstrate how to solve the POSG with reshaped rewards using algorithms available in the realm of MARL.
\subsection{Formulation of Equivalent POSG}\label{sec:POSG formulation}
It is straight forward to define a POSG whose state is identical with the team trajectory state $x_t$, i.e., $S = V_1 \times \cdots \times V_n \times 2^\Sigma$. However, by doing so, the satisfaction of TLTL specifications can only be determined based on the entire team trajectory. Hence, defining the reward function as the TLTL robustness contradicts the Markovian behaviors of the POSG in the sense that the reward $r$ only depends on current state of the system. Therefore, an extra element must be added to the POSG as a tracer of the whole team in the process of satisfying the global task. The state of the FSPA corresponding to the TLTL formula suits this job perfectly, and the state space of the POSG will be a product of the team trajectory states and the FSPA states. Now we are ready to introduce the following POSG. We use $v_t^i \in V_i$ and $a_t^i \in A_i$ to denote the state and action of agent $i$ at time $t$. 
\begin{definition}[FSPA augmented POSG] \label{def:FSPA-POSG}
    Given Problem \ref{problem formulation} and an FSPA $\mathcal{A} = \langle Q, q_0, F, T_{r}, \Psi, \mathcal{E},\\ b \rangle$ corresponding to the TLTL specification, an FSPA augmented POSG (FSPA-POSG) is defined as a tuple $\langle I, S, s_0, \{A_i, O_i, B_i\}_{i\in I}, p, \{r_i\}_{i\in I}\rangle$, in which $I=\{1,2,\ldots,n\}$ is the index set for agents, $S = V_1 \times \cdots \times V_n \times 2^\Sigma \times Q$ is the discrete state space.  $s_0 = ({v_0}_1, \ldots, {v_0}_n, \bar\Sigma_0, q_0)$ is the initial state. The action space $A_i = V \cup \Sigma_i \cup \epsilon$ of agent $i$ is identical with the action space of transition system $\mathcal T_i$. The observation space $O_i=V_i$ is the state space of $\mathcal T_i$. The observation function $B_i:S\rightarrow V_i$ maps the state of the FSPA-POSG to the state of agent $i$, which means that the policy of agent $i$ only depends on its own state. Let $s_t = (v_t^1, \ldots, v_t^n, \bar\Sigma_t, q_t)$ and $a_t = (a_t^1, \ldots, a_t^n)$ be the state of the FSPA-POSG and the joint action of all agents at time $t$ respectively. Then, the transition probability from $s_t$ to $s_{t+1}$ under action $a_t$ is defined as 
    \begin{equation*}
    p(s_{t+1}| s_t, a_t) = \\
        \begin{cases}
            1 & \text{if} \; \forall i \in I, v^i_t\xrightarrow{a_t^i}_{\mathcal{T}_i} v_{t+1}^i,\; \bar\Sigma_{t+1}=\{\cup_{i\in I}a_t^i | a_t^i\in\Sigma\}\ \text{and} \;\rho(x_t, \psi_{q_t,q_{t+1}}) > 0\\ 
            0 & \text{otherwise}
        \end{cases}
    \end{equation*}
    where $\psi_{q_t,q_{t+1}} = b(q_t,q_{t+1}) \in \Psi$ is the edge guard of the transition from $q_t$ to $q_{t+1}$ and $x_t$ is a state of the Team Trajectory included in $s_t$. Let the reward function $r_1=\ldots=r_n=r$ be defined as
    \begin{equation}
        r(s_t,a_t,s_{t+1}) = \\
        \begin{cases}
            C & \text{if} \; q_{t+1} \in F \; \text{and} \; q_t \not\in F\\
            -C & \text{if} \; q_{t+1} \in T_r\; \text{and} \; q_t \not\in T_r \\
            0 & \text{otherwise}
        \end{cases}
    \end{equation}\label{eqn:original reward}
    \vspace{-5pt}
    where $C>0$ is a constant.
\end{definition}

In this paper, we assume that the FSPA-POSG has a finite horizon $T$. The rationale of this assumption lies in the fact that any accepting or rejecting team trajectory must be finite (the corresponding FSPA reaches one state either in $F$ or in $Tr$). However, $T$ must be large enough to give agents adequate time to complete the task and thus it becomes a design factor of our approach. Clearly, since the team trajectory state is included in the FSPA-POSG state, constructing a team trajectory $x_{0:T}$ from $s_{0:T}$ is very simple, then constructing a TMS is also straightforward. 

\subsection{Temporal Logic Reward Shaping}\label{sec:TLRS}
As discussed in Sec. \ref{sec:intro}, we assume that the state of the FSPA-POSG is fully observable during training, while the observation function during deploying is as defined in Def. \ref{def:FSPA-POSG}. The reward function defined above only provides feedback to agents at the very end of each episode, which could cause serious problems when the probability of success under randomly selected actions converges to zero as the mission becomes more and more complicated. Missions described by temporal logics are expected to be complicated. Therefore, reward shaping techniques are introduced to add intermediate rewards to improve the speed and successful rate of learning.

We developed two additional rewards for each agent $i$ to guide learning: $r^i_\rho$ and $r^i_J$. $r^i_\rho$ is defined based on TLTL's quantitative semantics as follows:
\begin{equation}
\label{eqn:reward rho}
    r^i_\rho(s_t,a_t,s_{t+1}) = \rho(x_{t+1},D_{q_{t+1}}) - \rho(\hat{x}^i_{t+1}, D_{\hat q_{t+1}}),
\end{equation}
where $D_{q_{t+1}} = \bigvee_{(q_{t+1}, q_{t+2})\in \mathcal{E}, q_{t+2}\not\in T_r} b(q_{t+1}, q_{t+2})$ is the disjunction of all predicates of the outgoing edges from $q_{t+1}$ to other non-trapped states, $\hat{x}^i_{t+1} = (v^i_t \times^n_{j=1, j\neq i} v^j_{t+1}, \hat{\bar{\Sigma}}_{t+1})$ is a predicted FSPA-POSG state that all agents take actions and transition to the next state except agent $i$, $\hat q_{t+1}$ is the corresponding predicted automaton state. 
The robustness value $\rho(\cdot,D_{q_{t+1}})$ can be interpreted as a metric that measures how far it is from leaving the FSPA state $q_{t+1}$ given $x_{t+1}$. The individual reward $r^i_\rho$ is designed to be the difference between the robustness value under the whole team's actions and the robustness value with agent $i$ being idle. This is reasonable because of two reasons. First, $\hat{x}^i_{t+1}$ is generated by only using past observations so it does not require the agent model $\delta_i$. Second, since we know $\Sigma_i$, the value of predicate functions $do_i(\cdot)$ can be predicted by removing all the services that requires agent $i$'s action from the set of services that are finished at $t$, which makes $\hat{q}_{t+1}$ available. $r_\rho^i$ is able to distinguish each agent's contribution to the global behavior, and thus provide more informative rewards to agents.

By incorporating $r^i_\rho$, the agents are encouraged to leave the current FSPA state. However, the limitation of $r^i_{\rho}$ is that it cannot distinguish whether a transition is beneficial or not. For example, in Fig. \ref{fig:FSPA}, the bad transitions that we want to avoid are colored by orange. As for state $q_5$, two agents have already reached the location of $\sigma_3$ and the next thing to do is to finish $\sigma_3$ cooperatively. However, if they leave the vertex where $\sigma_3$ lives, $r^i_\rho$ will generate positive rewards because the FSPA has left $q_5$, which clearly violates our intentions. It is true that the learning can still converge to the correct policies, but only through the effect of discount factor $\gamma < 1$, in the sense that transition from $q_5$ to $q_4$ will delay the time of completing the mission and thus yield a lower final reward. 

To overcome this issue, we design another reshaped reward signal $r^i_J$, which can be seen as a "potential energy" of the corresponding FSPA state. Before giving the definition of $r^i_J$, we first introduce the concept of path length. Let $H(q,q')$ denote the set of all finite trajectories from $q \in Q$ to $q' \in Q$. The path length L is defined as
\begin{equation}
\vspace{-3pt}
\label{eqn:path length}
    L(\textbf{q}) = \sum_{k=1}^{n-1}\omega_{\mathcal{A}}(q_k, q_{k+1}),
    \vspace{-1pt}
\end{equation}
where $\textbf{q} = q_1\ldots q_n \in H(q_1, q_n)$ is a finite sequence of FSPA states connected by transitions in $\mathcal{E}$ and $\omega_{\mathcal{A}}: \mathcal{E} \to \mathbb{R}^+$ is a weight function. Then a distance function from $q$ to $q'$ can be defined as
\begin{equation}
\vspace{-2pt}
    \label{eqn:distance_q}
    d(q,q') = \begin{cases}
        \min_{\textbf{q} \in H(q,q')} L(\textbf{q}) \qquad & \text{if}\; H(q,q') \neq \emptyset \\
        \infty \qquad & \text{if}\; H(q,q') = \emptyset.
    \end{cases}
\vspace{-2pt}
\end{equation}
Now an energy function $J(q)$, $q \in Q$ can be defined as
\begin{equation}
\vspace{-2pt}
    \label{eqn:energy function}
    J(q) = \begin{cases}
        0 \qquad & \text{if}\; q \in F \\
        \min_{q'\in F} d(q,q') & \text{if}\; q \not\in F.
    \end{cases}
    \vspace{-2pt}
\end{equation}
In words, the energy function $J$ of a state $q$ is the minimum weighted sum of transitions it takes to reach the set of final states. Finally, $r^i_J$ is constructed as follows:
\begin{equation}
\vspace{-2pt}
    \label{eqn:rJ}
    r^1_J = \cdots = r^n_J = J(q) - J(q_{t+1}).
    \vspace{-2pt}
\end{equation}
$r^i_J$ ensures that agents will receive an assessment of every FSPA transition immediately. There is no standard procedure for determining the weight function $\omega_{\mathcal{A}}$ and hence it becomes a design factor.

\subsection{Solving the equivalent FSPA-POSG}\label{sec:Solving POSG}

We use the policy gradient method with policy graphs as in \cite{meuleau1999learning} and \cite{peshkin2001learning} to find the optimal policy for each agent in the FSPA-POSG. In the training phase, both the original reward and the two reshaped rewards require the information of the FSPA state, which is determined by all the agents. Hence, a centralized coordinator is necessary. We assume that the communication among all agents during the training phase is available, which is reasonable as stated in Sec. \ref{sec:intro}. After training, each agent obtains a policy that tracks its own history states to decide the action to be taken at each time step, i.e., the policies are distributed. 

\section{Simulation}\label{sec:simulation}
To verify the efficacy of the method described in Sec. \ref{sec:solution}, we conducted $500$ simulations of training on Example \ref{example} with reshaped rewards $r^i_\rho$, $r^i_\rho + r^i_J$ and $0$ (only original reward), respectively. In each simulation, we trained the policies for $10000$ episodes. We set the discount factor $\gamma$ to $0.995$, the learning step size $\alpha$ to $0.01$, the constant C in Eqn. \eqref{eqn:original reward} to $1$. The number of internal states of the policy graphs are chosen to be 10 for both agents. In the process of training policy graphs, we utilized Boltzmann exploration with a temperature parameter of 1. As a comparison, we also applied decentralized Q-learning \cite{matignon2012independent} with reshaped rewards $r^i_\rho + r^i_J$ and an $\epsilon$-greedy exploration strategy to learn policies without memory. 

\begin{figure}
\centering
\begin{minipage}[t]{0.55\textwidth}
\centering
\includegraphics[height=4cm]{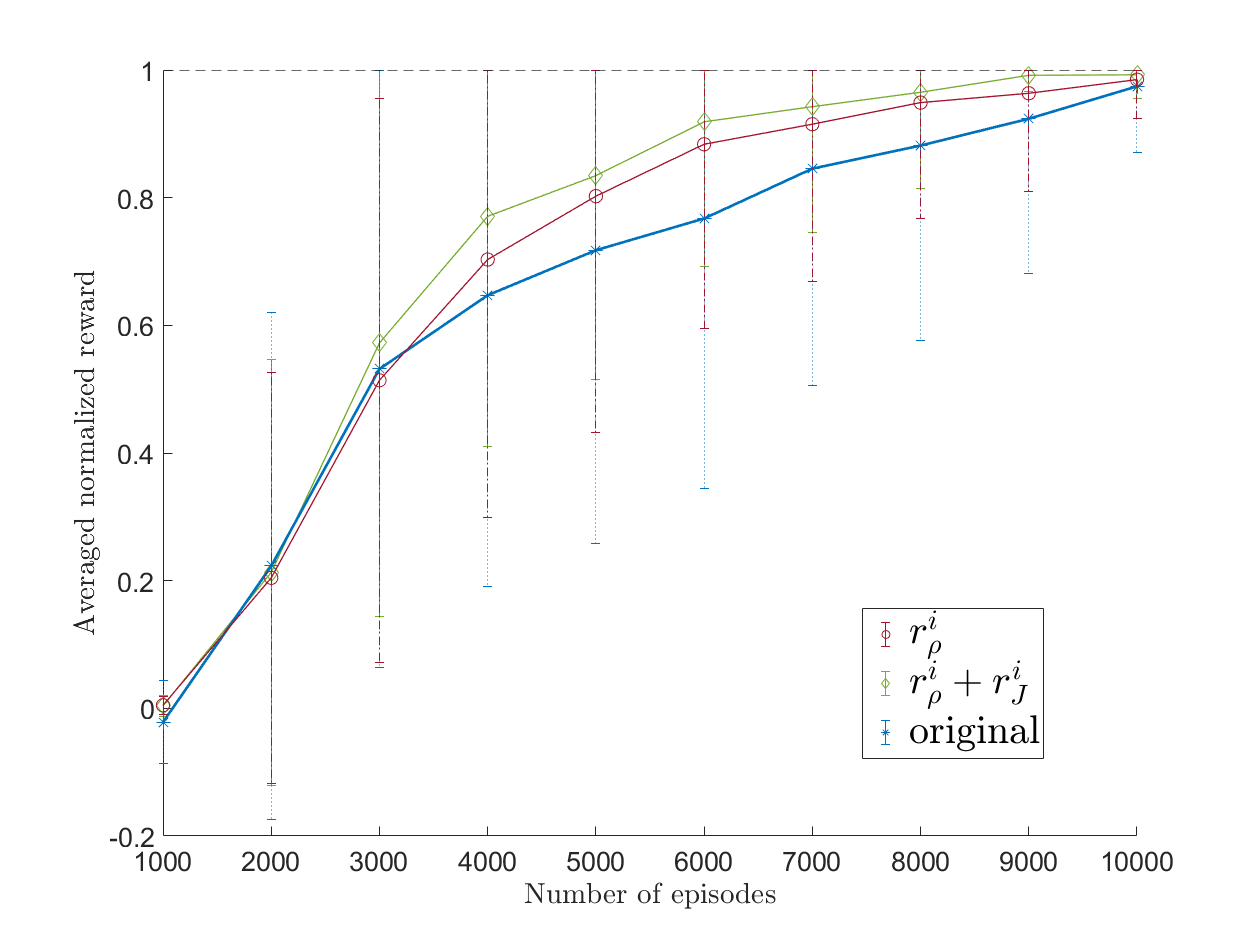}
\caption{\small The averaged learning curves of the three rewards. The values on the $y$-axis represent the normalized original reward signal in Eqn. \eqref{eqn:original reward}.}
\label{fig:L-curves}
\end{minipage}
\quad
\begin{minipage}[t]{0.41\textwidth}
\centering
\includegraphics[height=4cm]{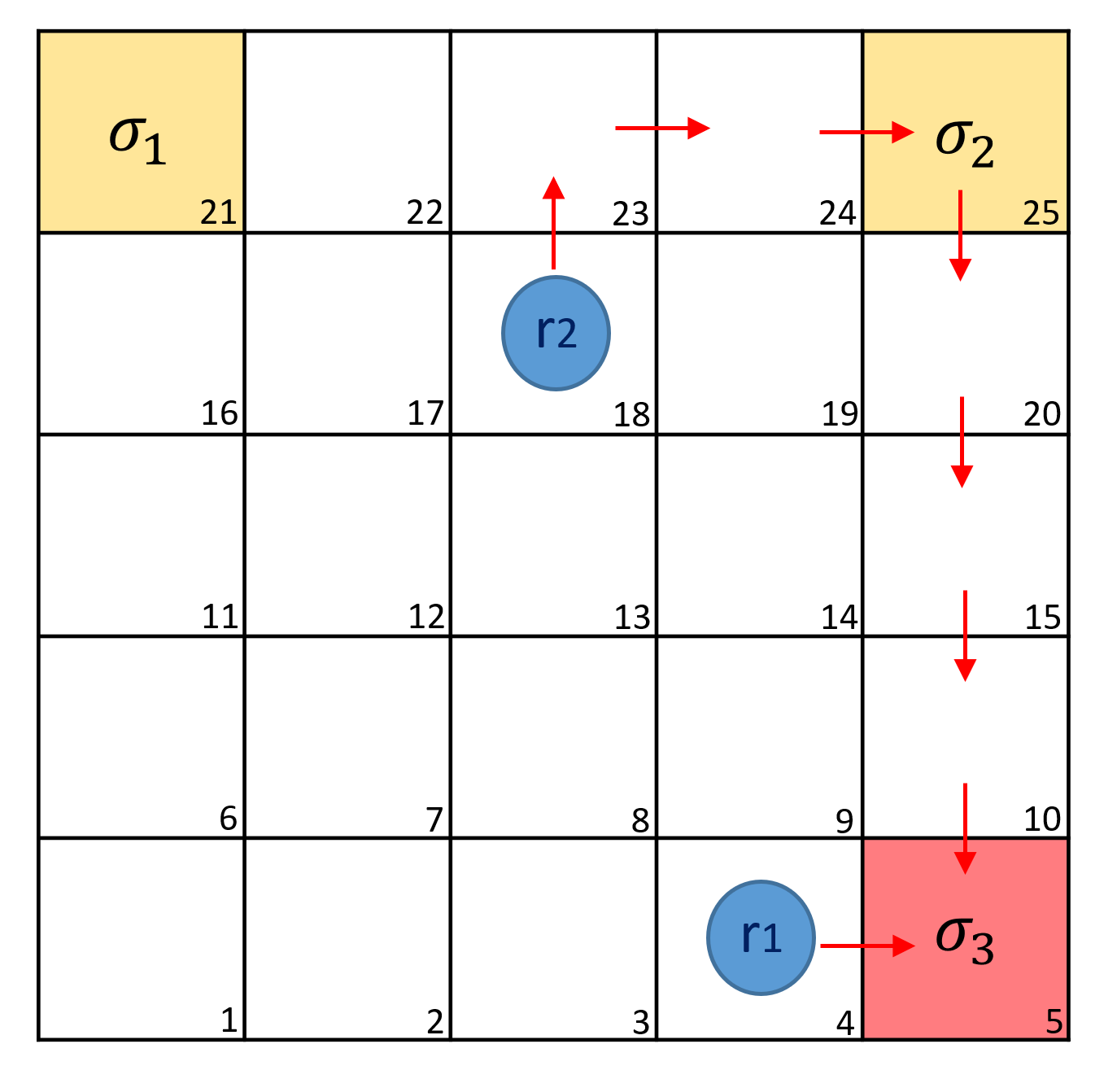}
\caption{\small Final trajectories generated by policy graphs after training. }
\label{fig:route}
\end{minipage}
\vspace{-10pt}
\end{figure}

It was not guaranteed that agents' policies would always converge to the optimal ones. The rates at which the optimal policies were learned within $10000$ episodes are shown in Table \ref{tb:1}. We can see that using $r_J^i+r_\rho^i$ boosted the convergence rate by $12.8\%$ compared to the original reward. Note that memoryless Q-learning completely fails the task. The averaged learning curves in which optimal policies successfully converged are shown in Fig. \ref{fig:L-curves}. We can see that the proposed reward shaping technique can accelerate the speed of learning and significantly reduce the variance. 

The behaviors of agents governed by the learned policies are visualized in Fig. \ref{fig:route}. In example \ref{example}, there are two possible routes to finish the TLTL task: finish $\sigma_1$ then $\sigma_3$; finish $\sigma_2$ then $\sigma_3$. It can be seen that the agents are able to figure out the optimal policy with minimum route length.

We also inspected the final policy graphs and their corresponding trajectories in simulations where agents with reshaped rewards failed to satisfy the TLTL specification. It turns out that in the majority of the failures, the agents 
would first finish either $\sigma_1$ or $\sigma_2$ and then never reach $\sigma_3$ by wondering among several vertices near $\sigma_1$ or $\sigma_2$. One possible cause is that the gradient ascent took steps that were too large, which cannot be fully handled by simply decreasing $\alpha$ \cite{schulman2017proximal}. 
More advanced policy gradient methods such as PPO \cite{schulman2017proximal} or TRPO \cite{schulman2015trust} might solve this issue. We will explore this direction in future research.

\begin{table}
\vspace{-10pt}
\centering
\setlength{\belowcaptionskip}{10pt}
\caption{Rates of convergence to the optimal policies with different rewards}
\label{tb:1}
\begin{tabular}{|l|c|c|c|c|}
\hline
Reshaped rewards& $r^i_\rho$ & $r^i_\rho+r^i_J$ & $0$ & $r^i_\rho+r^i_J$ (memoryless) \\
\hline
Optimal policy convergence rate & $68.4\%$& $83.0\%$ & $70.2\%$ & $0\%$\\
\hline
\end{tabular}
\vspace{-10pt}
\end{table}

\section{Conclusion} \label{sec:discussion}
In this paper, we applied MARL to synthesize distributed controls for a heterogeneous team of agents from Truncated Linear Temporal Logic (TLTL)  specifications. We assumed a partially observable environment, where each agent's policy depends on the history of its own states. We used the TLTL robustness as the reward of the MARL and introduce two additional reshaped rewards. Simulation results demonstrated the efficacy of our framework and showed that reward shaping significantly improves the convergence rate to the optimal policy and the learning speed. Future work includes consideration of non-deterministic SGs and other policy gradient methods.

\acks{This work was supported by NSF under Grant IIS-2024606 and Grant OIA-2020983.}

\bibliography{ref}

\end{document}